\documentclass{article}
\usepackage{ijcai20}

\usepackage{bbm}
\usepackage{times}
\usepackage{soul}
\usepackage{url}
\usepackage[hidelinks]{hyperref}
\usepackage[utf8]{inputenc}
\usepackage[small]{caption}
\usepackage{graphicx}
\usepackage{amsmath}
\usepackage{amsfonts} 
\usepackage{amsthm}
\usepackage{booktabs}
\usepackage{algorithm}
\usepackage{algorithmic}
\usepackage{subfigure}
\newtheorem{definition}{Definition}

\title{An interpretable neural network model through piecewise linear approximation}

\author{
Mengzhuo Guo$^{1,2}$\and
Qingpeng Zhang$^2$\footnote{Corresponding Author. E-mail:qingpeng.zhang@cityu.edu.hk}\and
Xiuwu Liao$^{1}$\and
Daniel Dajun Zeng$^{3,4}$\\
\affiliations
$^1$School of Management, Xi’an Jiaotong University\\
$^2$School of Data Science, City University of Hong Kong\\
$^3$The State Key Laboratory of Management and Control for Complex Systems, Institute of Automation, Chinese Academy of Sciences\\
$^4$Department of Management Information Systems, Eller College of Management, The University of Arizona
}

\begin{document}

\maketitle

\begin{abstract}
Most existing interpretable methods explain a black-box model in a \textit{post-hoc} manner, which uses simpler models or data analysis techniques to interpret the predictions after the  model is learned. However, they (a) may derive contradictory explanations on the same predictions given different methods and data samples, and (b) focus on using simpler models to provide higher \textit{descriptive} accuracy at the sacrifice of prediction accuracy. To address these issues, we propose a hybrid interpretable model that combines a piecewise linear component and a nonlinear component. The first component describes the explicit feature contributions by piecewise linear approximation to increase the expressiveness of the model. The other component uses a multi-layer perceptron to capture feature interactions and implicit nonlinearity, and increase the prediction performance. Different from the \textit{post-hoc} approaches, the interpretability is obtained once the model is learned in the form of feature shapes. We also provide a variant to explore higher-order interactions among features to demonstrate that the proposed model is flexible for adaptation. Experiments demonstrate that the proposed model can achieve good interpretability by describing feature shapes while maintaining state-of-the-art accuracy. 
\end{abstract}

\section{Introduction}

Recent research on interpretability explained the predictions in a \textit{post-hoc} manner: given a trained predictive model with predicted scores $F\left(\mathbf{x}\right)$, use extracted information to
explain how the model made predictions. Such extracted information can be analyzed and displayed by data analysis techniques, such as gradient-based methods \cite{tsang2017detecting,sundararajan2017axiomatic,shrikumar2017learning,yosinski2015understanding}, and sensitivity analysis \cite{ribeiro2016should,lundberg2017unified}, or by using a mimic model $\hat{F}$ to minimize $\Vert\hat{F}\left(\mathbf{x}\right)-F\left(\mathbf{x}\right)\Vert$, such as tree- and rule-based models \cite{li2019combining,che2016interpretable,letham2015interpretable,wang2019hybrid}. Summing up, the \textit{post-hoc} methods do not change or improve the underlying predictive model and use simpler forms, such as linear models, to explain the relationships in a complex predictive model, thereby making it easier for users to understand the interpretations. 

\textit{Post-hoc} methods, though effective in interpreting models in an easy-to-understand way, have two limitations: (a) When we use different \textit{post-hoc} methods to explain the same predictions, the explanations may be contradictory with each other \cite{alvarez2018towards}. The human cost for determining these methods and the samples to train the \textit{post-hoc} models are prohibitive with larger datasets. Moreover, it is unclear whether we can aggregate different \textit{post-hoc} explanations \cite{tan2018learning}; (b) The \textit{post-hoc} methods focus on using simpler models to provide a higher \textit{descriptive} accuracy, which measures the degree to which a \textit{post-hoc} method properly describes the patterns learned by a predictive model. Usually, the simplicity of \textit{post-hoc} methods helps users understand the patterns, but provides imperfect representations of nonlinear relationships among variables in complex black-box models. Their interpretability is obtained at the expense of prediction accuracy \cite{murdoch2019interpretable}. 

To overcome these limitations, there is a need for model-based interpretations, which come from the construction of the prediction model \cite{murdoch2019interpretable}. Such models (a) have a predefined structure and can readily describe relationships between input features and predictions \textit{once} the models are learned. It does not require the determination of the \textit{post-hoc} models and is estimated on all training data, therefore it avoids the interpretations being changed dramatically when different methods or subsets of the data are used. (b) Such models should interpret the relationships between input features and predictions in a simple and explicit way, such as describing feature contributions via providing feature shapes that can be easily understood by users. In the meanwhile, it should be expressive enough to properly fit the data and achieve good prediction performance.  

In this work, we propose a hybrid \textbf{Pi}ecewise \textbf{Li}near and \textbf{D}eep (\textit{PiLiD}) model under a Wide $\&$ Deep (short for W$\&$D) scheme \cite{cheng2016wide} as shown in Figure \ref{fig-prop}. The proposed model is comprised of a piecewise linear component and a nonlinear component. The first component uses piecewise linear functions to approximate the complex relationships between the input features and predictions. Such a form is explicit enough to describe the feature contributions by providing feature shapes, yet increases the expressiveness of the model. The other component uses a multi-layer perceptron (MLP) to capture the feature interactions and increase the prediction performance. The two components are jointly trained and the interpretability (in the form of piecewise linear functions) is obtained once the model is learned. This work has the following contributions:
\begin{itemize}
\item We propose the \textit{PiLiD} model that enhances the interpretability under a W$\&$D scheme. Different from the \textit{post-hoc} methods, the interpretability is obtained once the prediction model is learned. 
\item The proposed \textit{PiLiD} model is flexible for adaptations. We develop the \textit{PiLiB} model, a variant to extract interpretable higher-order interactions.
\item We integrate a predefined interpretable structure into the linear component to decipher the contribution of features by piecewise linear approximation. Such a form provides explicit feature shapes while preserving high expressiveness.
\item As a result of the joint-training scheme, we show that the model can describe complex feature shapes while improving model performance. 
\end{itemize}

\section{Related Work}
\subsection{Wide and Deep Scheme}
The Wide $\&$ Deep (W$\&$D) scheme jointly trains a linear model and a deep neural network to benefit from memorization and generalization \cite{cheng2016wide}.
% The input of linear part is usually transformed into a long vector to capture nonlinearity and all existing feature interactions in training data, whereas the input of deep neural network is the embedded low-dimensional and dense real-valued vector to explore new feature combinations that are not appeared in training data.
This framework takes advantage of both the Wide (linear model) and Deep (deep neural network) components by jointly training both together, and thus outperforms the models with either Wide or Deep component in a large-scale recommender system \cite{li2019combining,guo2018deepfm,zheng2017wide}. The Wide component of the W$\&$D scheme also provides interpretations of the features' main effect. Hence W$\&$D has been used to develop interpretable models. Tsang et al. \shortcite{tsang2017detecting} developed a model that modified the Deep component to only extract  pairwise interactions among features. Similarly, Kraus and Feuerriegel \shortcite{kraus2019forecasting} changed the Deep component to a recurrent neural network to capture the temporal information. Wang and Lin \cite{wang2019hybrid} proposed a hybrid model, in which an interpretable model and a black-box model function together to obtain a good trade-off between model transparency and performance. Effective as they are, these W$\&$D-based interpretable models' prediction performance is worse than the complex black-box model (the Deep part alone).

 %Summing up, the existing works have demonstrated that the W$\&$D framework can adapt for many mission-critical problems with better model performance or superior interpretability, however, all models mentioned above either focus on improving accuracy without considering model interpretability, or address model interpretability with sacrificing model accuracy. 

\subsection{Interpretable Methods}
There are mainly two categories of interpretable machine learning methods. The first refers to \textit{post-hoc} methods which initially learn an original predictive model and then explain how the model obtains the underlying predictions. This type of models usually fall into two sub-categories, prediction-level and dataset-level interpretations \cite{murdoch2019interpretable}.

The prediction-level methods focus on locally explaining why the model makes a particular prediction. They tell users what individual features or feature interactions are the most important by standard data analysis such as the gradient-based methods \cite{tsang2017detecting,sundararajan2017axiomatic,shrikumar2017learning,yosinski2015understanding}, the sensitivity analysis \cite{ribeiro2016should,lundberg2017unified}, the step-wise feature removal approaches \cite{schwab2019cxplain}, and the mimic models \cite{li2019combining,che2016interpretable,letham2015interpretable,wang2019hybrid}. In contrast, the dataset-level methods are interested in global explanations that explore more general relationships learned by the models. Tan et al. \shortcite{tan2018learning} described feature contributions by providing a global additive value function. Both prediction- and dataset-level methods suffer from the aforementioned drawbacks of \textit{post-hoc} scheme.

Another category of approaches predefines an interpretable structure, and provides insight into relationships between input features and predictions once the model is learned \cite{kraus2019forecasting}. %Kraus and Feuerriegel \shortcite{kraus2019forecasting} provided a structured-effect neural network for predictions of remaining useful life considering both model performance and interpretability, but they also describe the contribution of a feature by a simple coefficient.
Generalized additive model (GAM) can model extreme nonlinearity on individual features \cite{hastie1986generalized,lou2012intelligible}, but it cannot model feature interactions. In this regard, Lou et al. \shortcite{lou2013accurate} developed the GA$^2$M model, which first learns a base GAM without any interactions and then selects a number of pairwise interactions that minimize residuals. Unfortunately, the obtained individual and interacting feature contributions can be extremely complex because there is no regularization on their shapes, and this model is slow to converge. For this reason, Tsang et al. \shortcite{tsang2018neural} divided a neural network into several same-sized blocks and used $L_0$ regularization to model both uni-variate and high-order interactions, however, only a subset of individual features can be interpreted and the accuracy decreases because the model does not capture the nonlinearity other than the learned feature interactions. The main challenge, as the main purpose of this paper, is to come up with a structure that is simple enough to help users understand the rationale behind the predictions, while keeping the model sophisticated to take care of the non-interpretable nonlinearity in the data. 

%In contrast to the mentioned methods, the proposed framework uses a piecewise linear form for individual features as a prior and learns their shapes in the process of training the model. To the best of our knowledge, the proposed framework is the first one that is simultaneously (1) interpretable while maintaining high model accuracy; (2) able to provide feature shapes besides feature attributions; (3) flexible for adaptation.

\section{The Proposed \textit{PiLiD} Model}

\begin{figure*}
\centering
\includegraphics[scale=0.45]{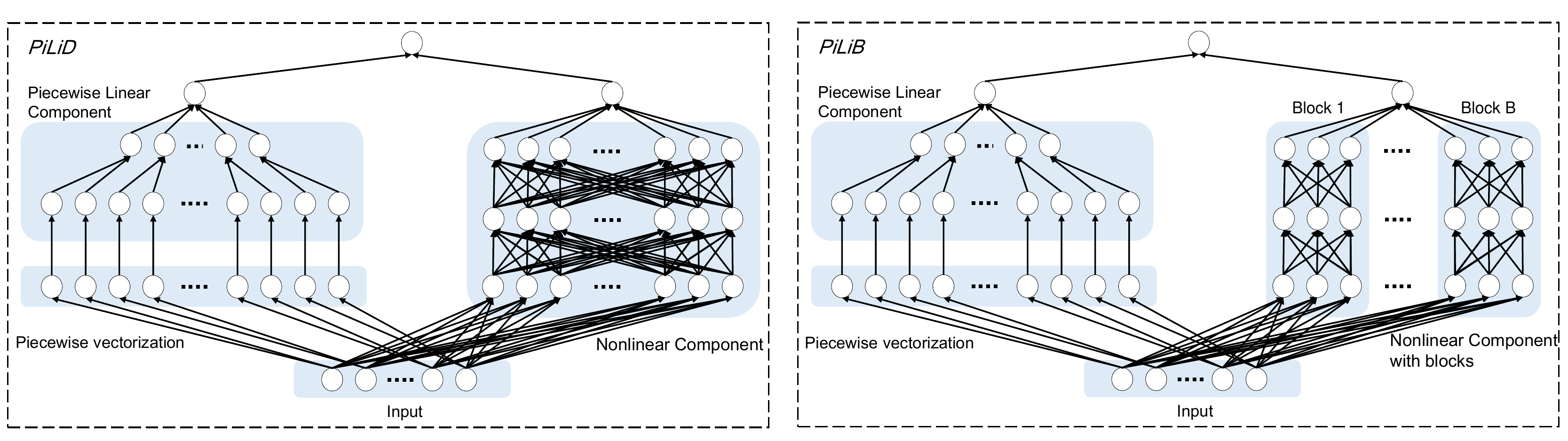}
\caption{The proposed \textit{PiLiD} (left) and \textit{PiLiB} (right) models. In \textit{PiLiB}, the whole neural network in the nonlinear component of \textit{PiLiD} is replaced with \textit{B} same-sized smaller neural networks (blocks).}
\label{fig-prop}
\end{figure*}
\subsection{Problem Setting}

Vectors are represented by boldface lower-case letters, for example $\textbf{x}$. Note that the index for $i$-th vector is in brackets, such as $\textbf{x}_{(i)}$ is the $i$-th data point, and the $j$-th element of a vector $\textbf{w}$ is $\textbf{w}_j$; matrices are denoted by boldface capital letters, for example $\textbf{W}$ and the element $(i, j)$ of $\textbf{W}$ is $W_{i,j}$.
Let $\{\textbf{x}_{(i)}, y_i \}_{i=1}^N$ be a set of $N$ data samples, where $\textbf{x}_{(i)} = (x_{i,1},\ldots,x_{i,j},\ldots, x_{i,m})^T \in \mathbb{R}^m$ is the $i$-th data point described by $m$ features, and $y_i$ is a real value (class label) to be predicted in regression (classification) problems. 

We assume that there exists a mapping function describing the relationship between the prediction and the individual features (main effects) and interacting features (interaction effects). To predict ${y_i}$, we define:
\begin{equation}
U\left({y_i}\right) = w_0 + \sum\limits_{j = 1}^m {u_j\left( {{x_{i,j}}} \right)}  + u\left( {{x_{i,1}}, \ldots {x_{i,m}}} \right) \label{eq-globalv1}
\end{equation}
where $w_0$ is a constant term\footnote{Although it can be left out in some prediction problems, for instance the binary classification problems where it does not affect the relative preference between two data samples. In this study, it can be decomposed into $m$ smaller values $w_{0,1},\ldots, w_{0,m}$, and each of them is set as a constant term in the corresponding marginal value function.}, $u_j\left(\cdot \right)$ is a \textit{marginal value function} of $j$-th feature, and $u\left(\cdot \right)$ is a function of all features. Eq.(\ref{eq-globalv1}) describes (a) a regression model if
$U(\cdot)$ is the identity, and (b) a classification model if $U(\cdot)$ is the logistic function
of the identity. 

Given the following definitions, we use piecewise linear functions to approximate the marginal value functions.

\begin{definition}
The characteristic points are some predefined values to partition the whole feature value scale into several pieces. \label{def-chara}
\end{definition}
For categorical features, the characteristic points ${\Psi_j} = \{ x_{i,j}|i=1,\ldots, N\}$ are all unique feature values. Let $\varphi_j^1,\ldots,\varphi_j^k,\ldots,\varphi_j^{{n_j}}$ be the ordered values of $\Psi_j,\varphi_j^k < \varphi_j^{k + 1},k = 1,\ldots,{n_j} - 1$, where ${n_j} = \left| {{\Psi_j}} \right|$ and ${n_j} \le N$. For numerical features, let $\left[\alpha_j, \beta_j \right]$, where $\alpha_j = min\{x_{i,j}|i=1,\ldots,N \}$ and $\beta_j = max\{x_{i,j}|i=1,\ldots,N \}$, be the whole evaluation value scale. We partition the scale into $\gamma_j$ equal sub-intervals $[\varphi^0_j, \varphi^1_j], [\varphi_j^1, \varphi_j^2 ],\dots,[\varphi_j^{\gamma_j-1}, \varphi_j^{\gamma_j} ]$, where $\varphi_j^k = {\alpha _j} + \frac{k}{{{\gamma _j}}}\left( {{\beta _j} - {\alpha _j}} \right),k = 0, 1, \ldots {\gamma _j}$ are characteristic points.

\begin{definition}
The feature vector $\boldsymbol{\Phi}_{(i)}\in \mathbb{R}^\gamma,\gamma={\sum\nolimits_{j = 1}^m {{\gamma _j}} }$ of data point $\textbf{x}_{(i)}$ is defined as follows:
\begin{equation}
\resizebox{.91\linewidth}{!}{$
\displaystyle
    \boldsymbol{\Phi}_{(i)} = {\left( {\underbrace {\phi_{i,1}^1, \ldots \phi_{i,1}^{{\gamma _1}},}_{1st } \ldots, \underbrace {\ldots, \phi_{i,j}^{k_j},  \ldots,}_{j-th } \ldots \underbrace {,\ldots \phi_{i,m}^{{\gamma _m} }}_{m-th }} \right)^T}
$}
\end{equation}
where $\phi_{i,j}^{k_j} = \left\{ \begin{array}{l}
1, \quad x_{i,j} > \varphi_j^{{k_j}},\\
\frac{{x_{i,j} - \varphi_j^{k_j-1}}}{{\varphi_j^{{k_j} } - \varphi_j^{k_j-1}}}, \: \varphi_j^{k_j-1} \le x_{i,j} \le \varphi_j^{{k_j}},\\
0,\quad \quad \text{otherwise}.
\end{array} \right. j=1,\dots,m,$ and ${k_j}=1,\dots,\gamma_j.$ \label{def-attr-vect}
\end{definition}

\begin{definition}
The marginal value vector $\textbf{u}$ is defined as:
\begin{equation}
    \textbf{u}={\left( {\Delta _1^1, \ldots ,\Delta _1^{{\gamma _1}}, \ldots ,\Delta _j^{{k_j}}, \ldots ,\Delta _m^1, \ldots ,\Delta _m^{{\gamma _m}}} \right)^T}
\end{equation}
where $\Delta_j^{k_j} = u_j(\varphi_j^{k_j})-u_j(\varphi_j^{k_j-1}),j=1,\dots,m,$ and $k_j = 1,\dots,\gamma_j$, is the difference of marginal values between two consecutive characteristic points.
\label{def-margin}
\end{definition}

We use an MLP to approximate the interaction effects $u\left( {{x_{i,1}}, \ldots {x_{i,m}}} \right)$. Here it is a feed-forward neural network with $L$ hidden layers. There are $p_l$ hidden units in the $l$-the layer and $p_0=m$. The layer matrices are denoted by $\textbf{W}^l \in \mathbb{R}^{p_l\times p_{l-1}}$, and bias vectors are denoted by $\textbf{b}^l \in \mathbb{R}^{p_l}, l=1,\ldots,L$. The non-linear activation function is denoted as $\phi(\cdot)$. Let $\textbf{w}^y \in \mathbb{R}^{p_L}$ and $b^y \in \mathbb{R}$ be the final weight vector and bias for the output $y$. In this way, the MLP with $L$ layers can be represented by:
\begin{align}
\textbf{h}^0 &= \textbf{x},\label{eq_mlp1}\\
\textbf{h}^l &= \phi \left( \textbf{W}^l \textbf{h}^{l-1} + \textbf{b}^l \right), l=1,\ldots,L,\label{eq_mlp2} \\
y &= \left(\textbf{w}^y \right)^T \textbf{h}^L + b^y.\label{eq_mlp3}
\end{align}

\subsection{Model Description and Training Process}

Given Definition \ref{def-margin} and Eq.(\ref{eq_mlp3}), the Eq.(\ref{eq-globalv1}) can be reformulated as follows:
\begin{equation}
U\left({y_i}\right) =w_0 + \textbf{u}^T \boldsymbol{\Phi}_i + u\left( {{x_{i,1}}, \ldots {x_{i,m}}} \right) \label{eq-globalv2}
\end{equation}

According to Definition \ref{def-attr-vect}, the vectorized input layer in piecewise linear component transforms the original data into a `wider' vector by piecewise linear partition. The input of each unit in this layer corresponds to an element in vector $\boldsymbol{\Phi}$, and every $\gamma_j$ units are decomposed from one feature. The first layer in the piecewise linear component has $\sum\nolimits_{j = 1}^m {{\gamma _j}}$ units and has a form $\textbf{y}=\textbf{w}^T\boldsymbol{\Phi}+\textbf{b}$, where $\textbf{w} = \left(w_1,\ldots,w_{\gamma} \right)^T$ and $\textbf{b} = \left(b_1,\ldots, b_{\gamma} \right)^T$. There is no any activation function. The next layer groups every $\gamma_j$ units ($m$ groups in total), and then sums the outputs of units in each group. Note that when $\gamma_j=1,j=1,\ldots,m$, the piecewise linear component \textit{PiLiD} degenerates to a simple linear model providing a single value describing feature contributions, i.e., feature attributions. When $\gamma_j\rightarrow \infty$, all feature values become characteristic points, \textit{PiLiD} has to fit the curve point-by-point. Thus, it can fit any curve given sufficient data. Without the constrain in $\gamma_j$, it may use an extremely complex shape to fit the curve, causing the over-fitting problem.

The outputs of two components are fed into a specific loss function for joint training. The joint training process uses the mini-batch stochastic optimization algorithm to back-propagate the gradients from both components at the same time. The general loss function is:
\begin{align}
\mathcal{R}\left( {\boldsymbol{\omega} ,\textbf{w},\textbf{b},\boldsymbol{\theta} } \right) &= \frac{1}{N}\left( {\sum\limits_{i = 1}^N {\mathcal{L}\left( {h\left( {\textbf{x}_{(i)} ;{\boldsymbol{\omega} ,\textbf{w},\textbf{b},\boldsymbol{\theta} } } \right),{y_i}} \right)} } \right) \nonumber \\
&+ \lambda \Omega \left( {\boldsymbol{\omega} ,\textbf{w},\textbf{b},\boldsymbol{\theta} } \right)
\label{eq-lossfunction1}
\end{align}
where $h(\cdot;{\boldsymbol{\omega} ,\textbf{w},\textbf{b}, \boldsymbol{\theta}})$ is a neural network with ${\boldsymbol{\omega} ,\textbf{w},\textbf{b},\boldsymbol{\theta}}$ parameters, ${\boldsymbol{\omega} ,\textbf{w},\textbf{b}}$ are the vectors of parameters in the piecewise linear component, ${\boldsymbol{\theta} }$ are the parameters, including weights matrices $\textbf{W}^l,l=1,\ldots,L$, bias vectors $\textbf{b}^l,l=1,\ldots,L$, weight vector $\textbf{w}^y$ and bias term $b^y$ for the output of nonlinear component, $\mathcal{L(\cdot)}$ is the loss function. Specifically,  $\mathcal{L(\cdot)}$ is the cross-entropy loss for classification problems and the mean square error for regression problems. $\lambda$ is a predefined coefficient for the regularization function $\Omega(\cdot)$ that can be $L_1$ or $L_2$ regularization.

The gradient-based algorithms can make the optimal solution robust and fast to converge by initializing the parameters in a neural network \cite{lecun2015deep}. In this study, the parameters in the piecewise linear component are not randomly initialized because their optimal values are supposed to indicate the contributions of individual features to the predictions. For this reason, we use the parameters obtained by a simple linear regression to initialize the parameters in the piecewise linear component. Such initialization forces the optimization algorithm starts from a more promising and interpretable solution region. As for the nonlinear component, we use a standard Gaussian initialization. The pseudo code for learning the proposed model is presented in Algorithm \ref{alg-trainprocess1} (for brevity, we raise a regression problem as an example):
\begin{algorithm}
\caption{Training process for the proposed \textit{PiLiD}.}
\label{alg-trainprocess1}
\begin{algorithmic}[1]
\REQUIRE Training data $\{\textbf{x}_{(i)}, y_i\}^N_{i=1}$, structure of nonlinear component $L$, predefined number of sub-intervals $\gamma_j,j=1,\dots,m$, type of loss function $\mathcal{L}(\cdot)$, regularization term $\lambda$ and regularization type $\Omega(\cdot)$, initialization coefficient $\sigma$ for nonlinear component, number of epoch and Batch size.\\
\ENSURE Parameters ${\boldsymbol{\omega} ^*},{\textbf{w}^*},{\textbf{b}^*},{\boldsymbol{\theta} ^*} = \mathop {\arg \min }\limits_{\boldsymbol{\omega} ,\textbf{w},\textbf{b}, \boldsymbol{\theta}} \{ \mathcal{R}({\boldsymbol{\omega} ,\textbf{w},\textbf{b}, \boldsymbol{\theta}})\}$. \\
\STATE Vectorize the feature values in $\textbf{x}_{(i)}$ into $\boldsymbol\Phi_{(i)}$ according to Definition \ref{def-attr-vect}. Let $\boldsymbol\Phi=(\boldsymbol\Phi_{(1)},\ldots,\boldsymbol\Phi_{(N)})^T$ and $\textbf{y}=(y_1,\ldots,y_N)^T$.\\
\STATE Initialize $\boldsymbol\omega \leftarrow (1,\ldots,1)^T$, $\textbf{w} \leftarrow \left(\boldsymbol\Phi\boldsymbol\Phi^T\right)^{-1}\boldsymbol\Phi\textbf{y}$, $\textbf{b}\leftarrow(0,\ldots,0)^T$, $\textbf{W}^l\leftarrow \mathcal{N}\left(0, \sigma \right)$, and $\textbf{b}^l\leftarrow(0,\ldots,0)^T,l=1,\ldots,L$.\\
\STATE Standard mini-batch stochastic gradient optimization algorithm.
\end{algorithmic}
\end{algorithm}

\subsection{A Variant}
To explore interacting effects in nonlinear component and demonstrate the flexibility of the proposed model, we substitute the nonlinear component with the state-of-the-art \textit{Neural Interaction Transparency (NIT)} model to decipher the high-order feature interactions \cite{tsang2018neural}.

The original \textit{NIT} model partitions a whole feed-forward neural network into several `blocks' with the same size. It uses the $L_0$ regularization to force some hidden units to have zero weights in each `block', and adds an extra term $\mathcal{L_K}$ in the loss function to encourage the model to learn smaller-sized interactions. We refer interested readers to Tsang et al.\shortcite{tsang2018neural}. In this work, we propose a variant of \textit{PiLiD}, namely the Piecewise Linear and Blocks (\textit{PiLiB}) to model all individual features and important interacting features. 

The piecewise linear component in \textit{PiLiB} is same as that in \textit{PiLiD}. The nonlinear component is divided into $B$ same-sized block (shown on the right in Figure \ref{fig-prop}). The loss function of \textit{PiLiD} is:
%\begin{equation}
%\mathcal{R} = \frac{1}{N}
%\left( {\sum\limits_{i = 1}^N {\mathcal{L}\left( {h\left( {\textbf{x}_{(i)} ;{\boldsymbol{\omega} ,\textbf{w},\textbf{b},{\tilde{\textbf{W}}^1 \odot \mathcal{T}\left( \textbf{G} \left( \boldsymbol\Phi \right) \right) } } } \right), \{\textbf{W}^l\}_{l=2}^{L+1}, \{\textbf{b}^l \}_{l=2}^{L+1},{y_i}} \right)} } \right) 
%+ \mathcal{L_K}'
%\end{equation}
\begin{align}
&\mathcal{R}_{PiLiB} =  \mathcal{L_K}' + \frac{1}{N}\times (\sum\limits_{i = 1}^N {}\mathcal{L}( h( \textbf{x}_{(i)} ;\boldsymbol{\omega} ,\textbf{w},\textbf{b},\tilde{\textbf{W}}^1 \odot \nonumber \\
&\quad \quad \quad \quad \quad \mathcal{T}( \textbf{G} ( \boldsymbol\Phi ) )    ), \{\textbf{W}^l\}_{l=2}^{L+1}, \{\textbf{b}^l \}_{l=2}^{L+1},{y_i} )  )  \\
&\mathcal{L_K}' = \max \left\{ {\left( {\mathop {\max }\limits_i \hat{k}_i} \right) - K,0} \right\}\nonumber \\
&\quad \quad \quad \quad \quad + \lambda_0{{\sum\limits_{i = 1}^B {\left( {{{\hat k}_i} - 2} \right)} } \mathord{\left/
 {\vphantom {{\sum\limits_{i = 1}^B {\left( {{{\hat k}_i} - 2} \right)} } {\sum\nolimits_{i = 1}^B {1\left( {k \ne 0} \right)} }}} \right.
 \kern-\nulldelimiterspace} {\sum\nolimits_{i = 1}^B {\mathbbm{1}\left( {\hat{k}_i \ne 0} \right)} }}
\end{align}
% %where $\mathcal{L_K}' = \max \left\{ {\left( {\mathop {\max }\limits_i \hat{k}_i} \right) - K,0} \right\} + \lambda_0{{\sum\limits_{i = 1}^B {\left( {{{\hat k}_i} - 2} \right)} } \mathord{\left/
%  {\vphantom {{\sum\limits_{i = 1}^B {\left( {{{\hat k}_i} - 2} \right)} } {\sum\nolimits_{i = 1}^B {1\left( {k \ne 0} \right)} }}} \right.
%  \kern-\nulldelimiterspace} {\sum\nolimits_{i = 1}^B {\mathbbm{1}\left( {k \ne 0} \right)} }}$.
Note that $\textbf{G}\approx1$ if a feature is active in a block, thus the estimated interaction order is defined as $\hat{k}_i = \sum\limits_{j=1}^m {G_{ij}(\Phi_{ij})},\forall i=1,\ldots,B$. The first term in $\mathcal{L_K}'$ limits the maximum interaction order to be equal to or less than $K$. Different with Tsang et al. \shortcite{tsang2018neural}, the second term encourages the model to learn pairwise interactions. In this way, we push the modelling of the main effects to the piecewise linear component and the pairwise interactions to the nonlinear component. The algorithm for training parameters in \textit{PiLiB} is in supplementary materials.

\begin{table*}
  \centering
  \caption{Average $MSE$ for regression problems when $N=20000$ (The best results are marked in bold. The number following \textit{PiLiD} is the predefined number of intervals for piecewise linear functions.).}
    \begin{tabular}{lrrrrr}
          \toprule
       \# features   & {10} & {20} & {30} & {40} & {50} \\
          \midrule
    \textit{PiLiD-1} & 0.2677 $\pm$ 0.0073 & 0.3041 $\pm$ 0.0147 & 0.2770 $\pm$ 0.0060 & 0.3406 $\pm$ 0.0139 & 0.3522 $\pm$ 0.0121 \\
    \textit{PiLiD-5} & 0.2666 $\pm$ 0.0070 & 0.3042 $\pm$ 0.0079 & \textbf{0.2752 $\pm$ 0.0067} & 0.3318 $\pm$ 0.0155 & 0.3553 $\pm$ 0.0152 \\
    \textit{PiLiD-10} & 0.2678 $\pm$ 0.0089 & 0.3060 $\pm$ 0.0158 & 0.2793 $\pm$ 0.0141 & 0.3383 $\pm$ 0.0184 & 0.3491 $\pm$ 0.0178 \\
    \textit{PiLiD-15} & 0.2662 $\pm$ 0.0070 & 0.3039 $\pm$ 0.0124 & {0.2760 $\pm$ 0.0087} & \textbf{0.3291 $\pm$ 0.0133} & 0.3473 $\pm$ 0.0134 \\
    \textit{PiLiD-20} & \textbf{0.2657 $\pm$ 0.0058} & \textbf{0.3023 $\pm$ 0.0077} & {0.2793 $\pm$ 0.0064} & 0.3296 $\pm$ 0.0129 & \textbf{0.3448 $\pm$ 0.0144} \\
    MLP   & 0.2705 $\pm$ 0.0133 & 0.3092 $\pm$ 0.0067 & 0.2816 $\pm$ 0.0035 & 0.3599 $\pm$ 0.0218 & 0.3681 $\pm$ 0.0236 \\
    SVM (rbf kernel)  & 0.2801 $\pm$ 0.0047 & 0.3262 $\pm$ 0.0079 & 0.2856 $\pm$ 0.0058 & 0.3935 $\pm$ 0.0083 & 0.3865 $\pm$ 0.0073 \\
    Random Forest    & 0.2703 $\pm$ 0.0052 & 0.3131 $\pm$ 0.0121 & 0.3160 $\pm$ 0.0069 & 0.4032 $\pm$ 0.0081 & 0.3998 $\pm$ 0.0083 \\
    Linear Regression    & 0.3518 $\pm$ 0.0053 & 0.3952 $\pm$ 0.0095 & 0.2873 $\pm$ 0.0058 & 0.3925 $\pm$ 0.0082 & 0.4176 $\pm$ 0.0098 \\
    GAM   & 0.2976 $\pm$ 0.0040 & 0.3250 $\pm$ 0.0060 & 0.2804 $\pm$ 0.0050 & 0.3808 $\pm$ 0.0052 & 0.3726 $\pm$ 0.0068 \\
    \bottomrule
    \end{tabular}%
  \label{tab-n20000}%
\end{table*}%

\begin{table*}
  \centering
  \caption{Averaged $MSE$ given proposed initialization and Gaussian initialization ($^*,^{**}$ are significantly different at 10\% and 5\% levels.).}
    \begin{tabular}{rrrrrrrrrrr}
    \toprule
          & \multicolumn{2}{c}{$\gamma_j=1$} & \multicolumn{2}{c}{$\gamma_j=5$} & \multicolumn{2}{c}{$\gamma_j=10$} & \multicolumn{2}{c}{$\gamma_j=15$} & \multicolumn{2}{c}{$\gamma_j=20$} \\
          \midrule
         \# objects & \multicolumn{1}{l}{Initi.} & \multicolumn{1}{l}{G. Ini.} & \multicolumn{1}{l}{Initi.} & \multicolumn{1}{l}{G. Ini.} & \multicolumn{1}{l}{Initi.} & \multicolumn{1}{l}{G. Ini.} & \multicolumn{1}{l}{Initi.} & \multicolumn{1}{l}{G. Ini.} & \multicolumn{1}{l}{Initi.} & \multicolumn{1}{l}{G. Ini.} \\ \midrule
    5000  & \textbf{0.3755}$^{**}$ & 0.3860 & 0.3704 & 0.3625 & 0.3752 & 0.3909 & 0.3684 & 0.3603 & 0.3686 & 0.3666 \\
    10000 & \textbf{0.3427}$^{**}$ & 0.3612 & 0.3407 & 0.3305 & \textbf{0.3348}$^{**}$ & 0.3443 & \textbf{0.3359}$^{**}$ & 0.3477 & \textbf{0.3341}$^{**}$ & 0.3428 \\
    15000 & 0.3182 & 0.3105 & \textbf{0.3147}$^{**}$ & 0.3251 & \textbf{0.3101}$^*$ & 0.3155 & \textbf{0.3105}$^*$ & 0.3176 & \textbf{0.3087}$^{**}$ & 0.3275 \\
    20000 & \textbf{0.3083}$^{**}$ & 0.3273 & \textbf{0.3066} & 0.3091 & \textbf{0.3081}$^{**}$ & 0.3209 & \textbf{0.3045} & 0.3091 & \textbf{0.3044} & 0.3055 \\ \bottomrule
    \end{tabular}%
  \label{tab-initial}%
\end{table*}%

\section{Experiments}

Our experiments are conducted on synthetic and real datasets. The simulations aim to answer the following questions: (a) Can the proposed \textit{PiLiD} accurately describe the actual marginal value functions while outperforming other predictive models? (b) Does the initialization process help improve interpretability and accuracy? The experiments on real datasets aim to show the advantages of \textit{PiLiD} and \textit{PiLiB} over several state-of-the-art interpretable methods.

\subsection{Simulations}

The synthetic data is generated as follows: (a) Randomly determine the coefficients of $m$ polynomial functions (in 10 degrees) as the actual marginal value functions. (b) Randomly determine $N$ objects and their feature values from interval $[0,1]$. (c) Use actual marginal value functions to calculate the marginal values of $N$ objects and randomly add some feature interactions. (d) For regression problem, $y$ is the summation of $m$ marginal values and selected interactions; For binary classification problem, we use a logistic function to assign 0 and 1 labels. In this section, we present the results of regression problems due to length limit. The classification problem has similar results, which are presented in the supplementary materials.

The MLP used in \textit{PiLiD} has a 100-200-400-400-200-100-1 structure. We use ADAM with 0.005 learning rate as the standard optimizer in Algorithm \ref{alg-trainprocess1}. For each problem setting, we run 20 trials. 80\% is selected as the training set and 20\% as the testing set. We present experimental results of regression problems with $N=20000$ in Table \ref{tab-n20000}. The full simulation results can refer to supplementary materials.

When the synthetic data size is small, both MLP and \textit{PiLiD} perform worse than classic statistical learning models (see results in supplementary materials when $N=5000$), such as SVM and Random Forest. When the data size is larger, more data are fed into the neural network, as a result, both MLP and \textit{PiLiD} perform better than statistical learning models (Table \ref{tab-n20000}). Encouragingly, we find that \textit{PiLiD} outperforms MLP in most tasks given the same neural network setting and the number of iterations. At last, as the increase in the number of predefined intervals $\gamma_j$, the piecewise linear component becomes more complex, and in turn increases the model accuracy. This indicates that the piecewise linear component could actually improve the performance if the learned interpretability is close to the actual one. Nevertheless, we suggest setting $\gamma_j=5,10,15$ to avoid the overfitting problem and additional computational cost.

\begin{table*}
  \centering
  \caption{Predictive performance of the proposed \textit{PiLiD} and \textit{PiLiB}. In both models, $\gamma_j=5$. In \textit{NIT} and \textit{PiLiB}, $K=3,B=20$.}
    \begin{tabular}{lrrrr}
    \toprule
          & bike sharing (MSE) & bank marketing (AUC) & spambase (AUC) & skill (AUC) \\
          \midrule
    \textit{PiLiD} & \textbf{0.068$\pm$0.0087} & \textbf{0.938$\pm$0.0031} & \textbf{0.978$\pm$0.0021} & \textbf{0.877$\pm$0.0058} \\
    \textit{PiLiB}& 0.107$\pm$0.0109 & 0.924$\pm$0.0028 & 0.965$\pm$0.0047 & 0.862$\pm$0.0047 \\
    MLP   & 0.071$\pm$0.0091 & 0.929$\pm$0.0027 & 0.966$\pm$0.0031 & 0.864$\pm$0.0049 \\
    \textit{NIT} & 0.090$\pm$0.0088 & 0.894$\pm$0.0367 & 0.961$\pm$0.0039 & 0.858$\pm$0.0051 \\
    GA & 0.137$\pm$0.0023 & 0.882$\pm$0.0098 & 0.935$\pm$0.0023 & 0.851$\pm$0.0037 \\
    GAM   & 0.215$\pm$0.0015 & 0.856$\pm$0.0032 & 0.923$\pm$0.0009 & 0.811$\pm$0.0023 \\
    GA$^2$M  & 0.101$\pm$0.0021 & 0.902$\pm$0.0055 & 0.959$\pm$0.0013 & 0.856$\pm$0.0038 \\
    \bottomrule
    \end{tabular}% 
  \label{tab-predictive}%
\end{table*}%

\begin{figure}
\centering
\subfigure[The proposed initialization.]{
\begin{minipage}[t]{0.5\linewidth}
\centering
\includegraphics[scale=0.25]{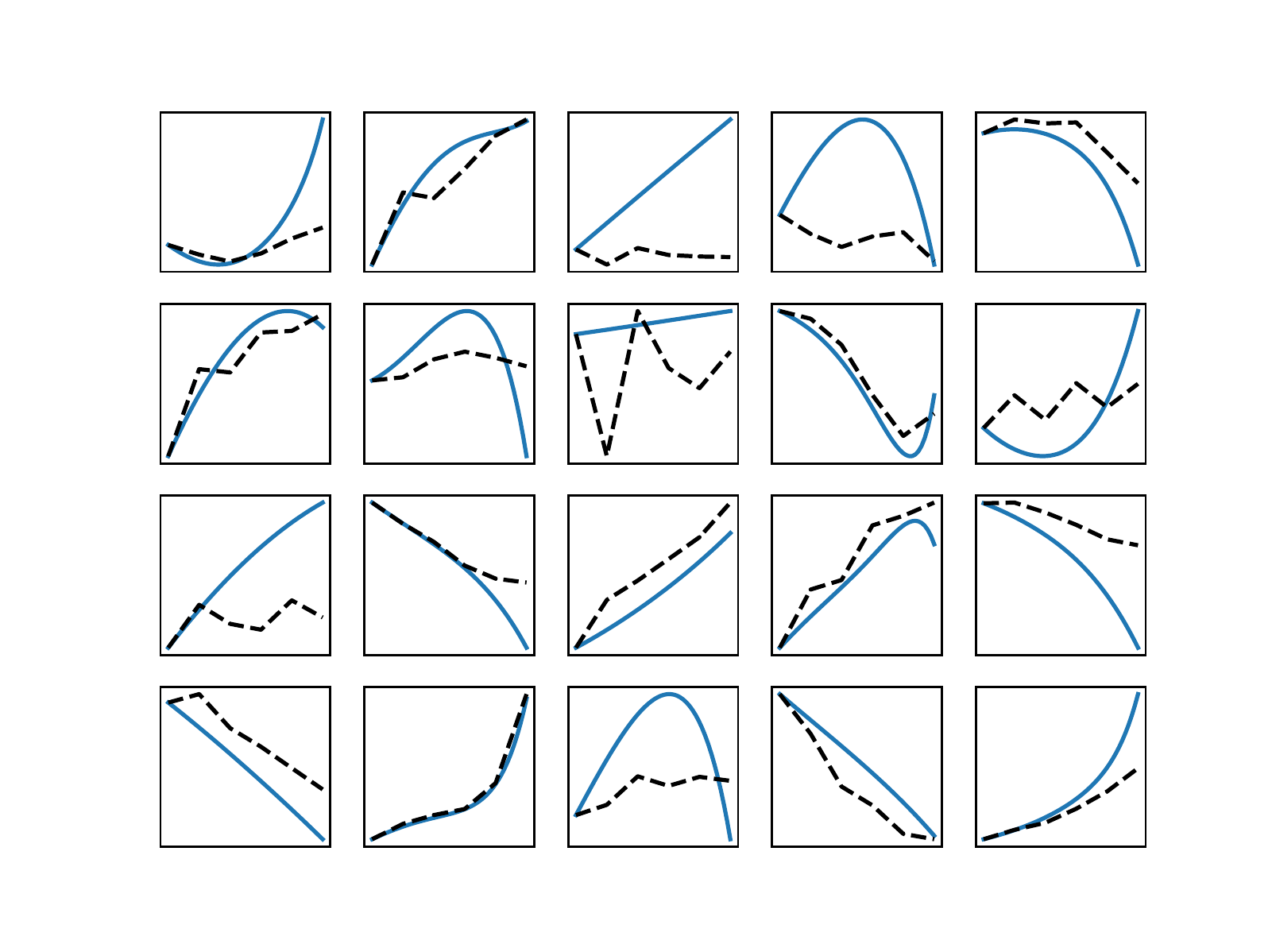}
%\caption{fig1}
\end{minipage}%
}%
\subfigure[The Gaussian initialization.]{
\begin{minipage}[t]{0.5\linewidth}
\centering
\includegraphics[scale=0.25]{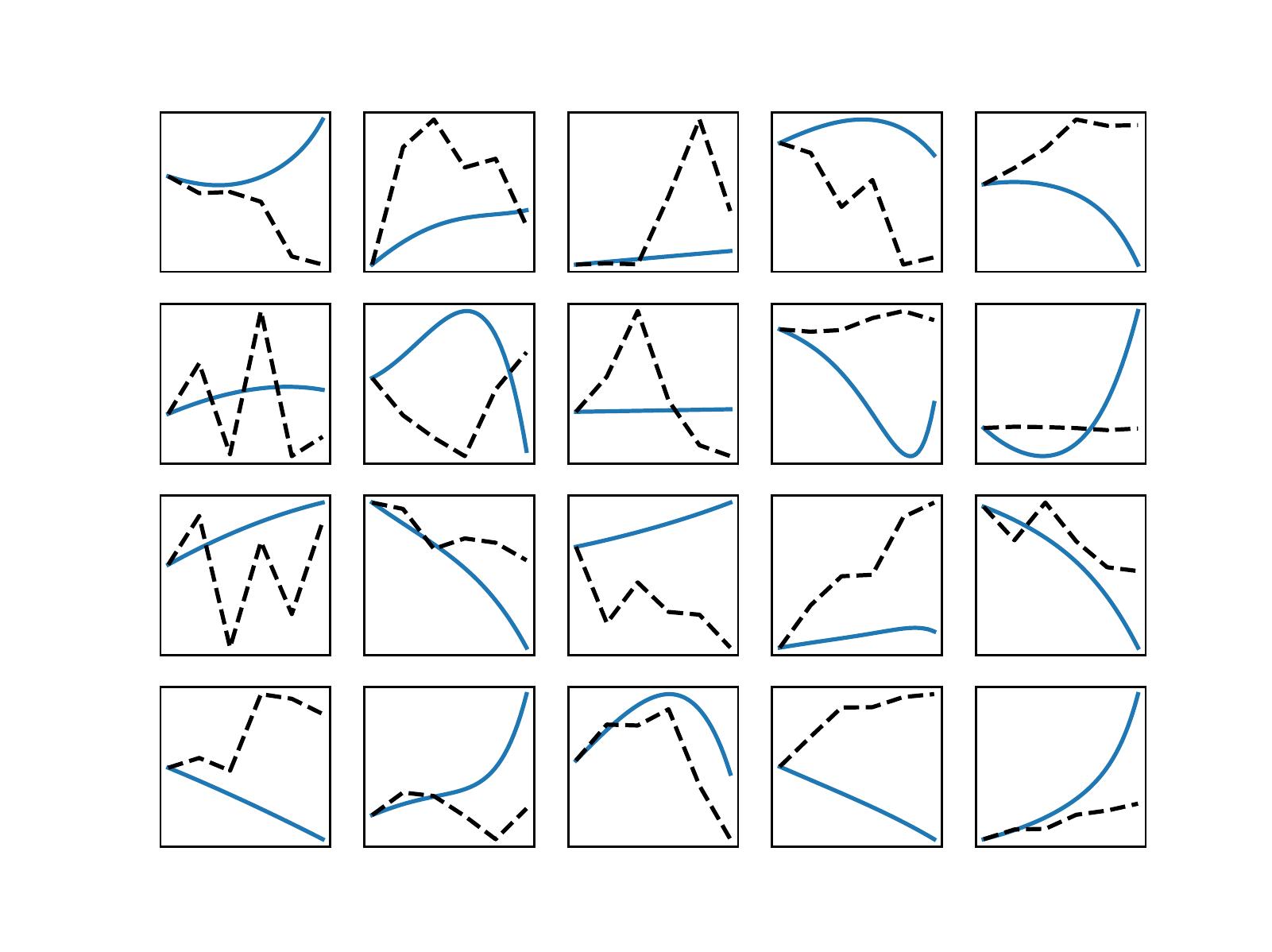}
%\caption{fig2}
\end{minipage}%
}%
\caption{An example of the learned marginal value functions given different initializations when $m=20, N=15000, \gamma_j=5$. All curves are averaged over 20 trials. The solid blue curves are the actual functions and the dashed black curves are learned functions.}
\label{fig-simulation}
\end{figure}

Summing up, the proposed \textit{PiLiD} performs better because (a) the piecewise linear form increases model expressiveness; and (b) the proposed initialization in Algorithm \ref{alg-trainprocess1} and the joint training process help obtain optimal solutions (see results in Table \ref{tab-initial}). To demonstrate that the proposed initialization helps obtain rational interpretations, we provide an example of learned marginal value functions by two initializations in Figure \ref{fig-simulation}. Obviously, although Gaussian initialization can sometimes obtain results as accurate as the proposed initialization, it fails to properly describe the actual marginal value functions.

\subsection{Real Data}

\begin{table}
  \centering
  \caption{Description on datasets.}
    \begin{tabular}{lllll}
    \toprule
          & Bike share & Bank market & Spambase & Skill \\
          \midrule
\# objects & 17,379 & 40,787 & 4,601 & 3,302\\
\# features & 15 & 48 & 57 & 18 \\
    \bottomrule
    \end{tabular}%
  \label{tab-descri}%
\end{table}%

We conduct experiments using four real datasets to compare the prediction performance of the proposed \textit{PiLiD} and \textbf{PiLiB}, and a set of baseline models including \textit{global additive explanation model} (GA) \cite{tan2018learning}, GAM \cite{lou2012intelligible}, GA$^2$M \cite{lou2013accurate}, and \textit{NIT} \cite{tsang2018neural} models. The datasets are described in Table \ref{tab-descri}\footnote{The datasets can be downloaded from supplementary materials.}.

Table \ref{tab-predictive} presents the performance of different models. Comparing with \textit{NIT} and MLP, the proposed \textit{PiLiD} performs better because it benefits from the extra piecewise linear component that considers more complex feature shapes and the jointly training process. Comparing with the GAM family, although GAMs can model extremely complex feature shapes, they cannot account for possible higher-order interactions (except for GA$^2$M, which only deals with the pairwise interactions). Therefore, both \textit{PiLiD} and \textit{PiLiB} outperform all GAMs. 

\begin{figure}
\centering
\subfigure[$\gamma_j=1$]{
\begin{minipage}[t]{0.3\linewidth}
\centering
\includegraphics[scale=0.15]{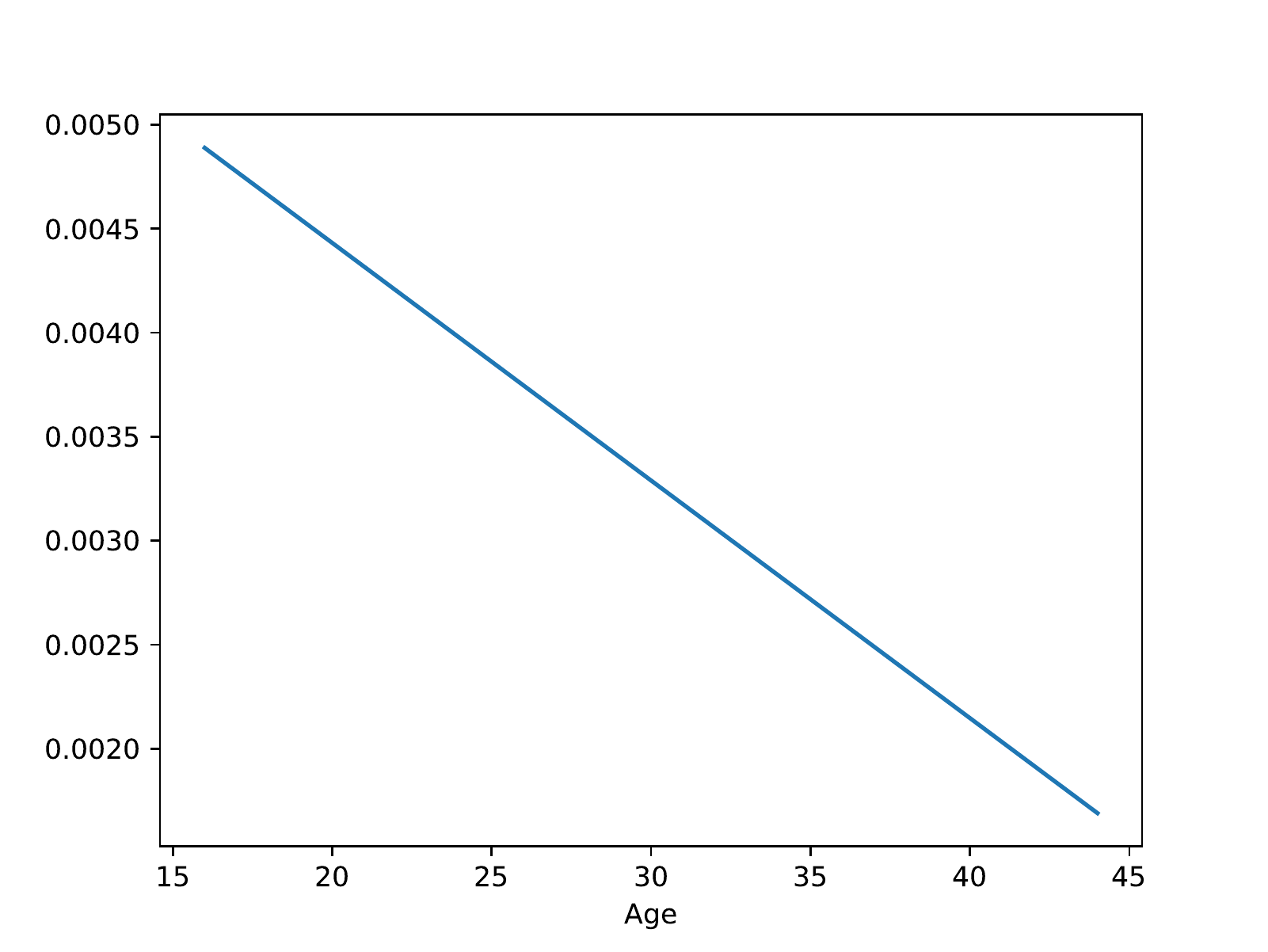}
%\caption{fig1}
\end{minipage}%
}%
\subfigure[$\gamma_j=5$]{
\begin{minipage}[t]{0.3\linewidth}
\centering
\includegraphics[scale=0.15]{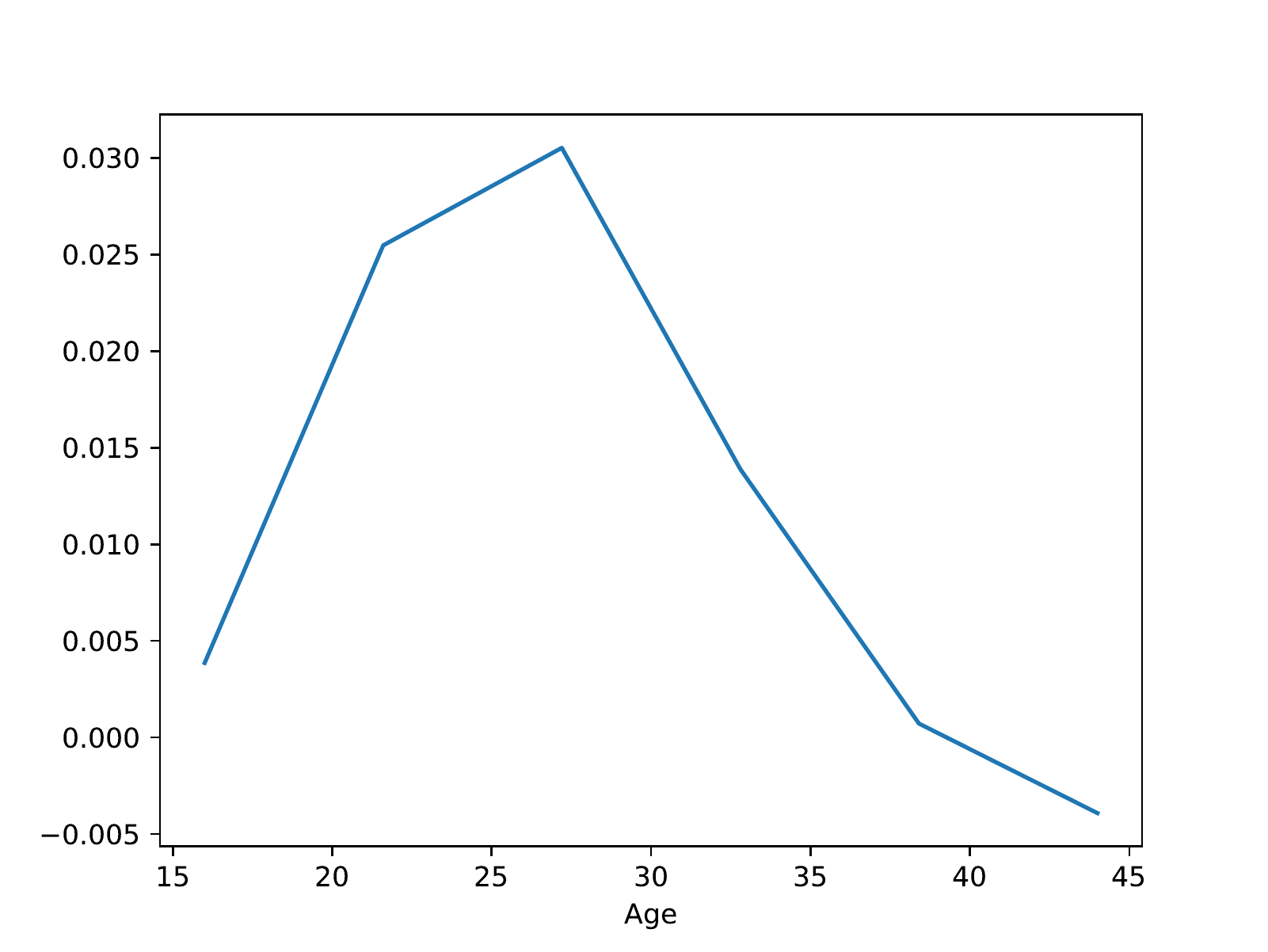}
%\caption{fig2}
\end{minipage}%
}%
\subfigure[$\gamma_j=10$]{
\begin{minipage}[t]{0.3\linewidth}
\centering
\includegraphics[scale=0.15]{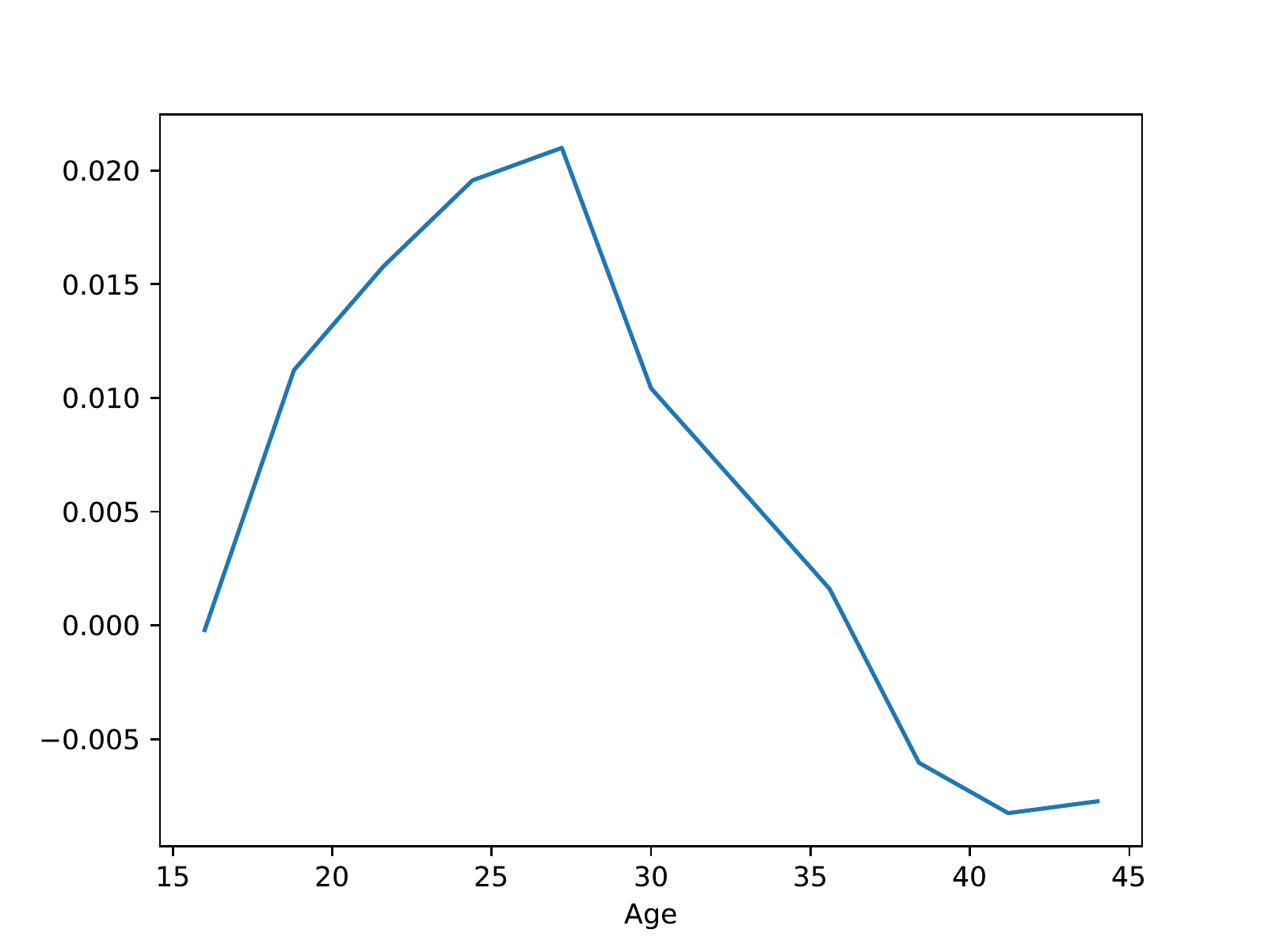}
%\caption{fig2}
\end{minipage}%
}%

\subfigure[$\gamma_j=15$.]{
\begin{minipage}[t]{0.3\linewidth}
\centering
\includegraphics[scale=0.15]{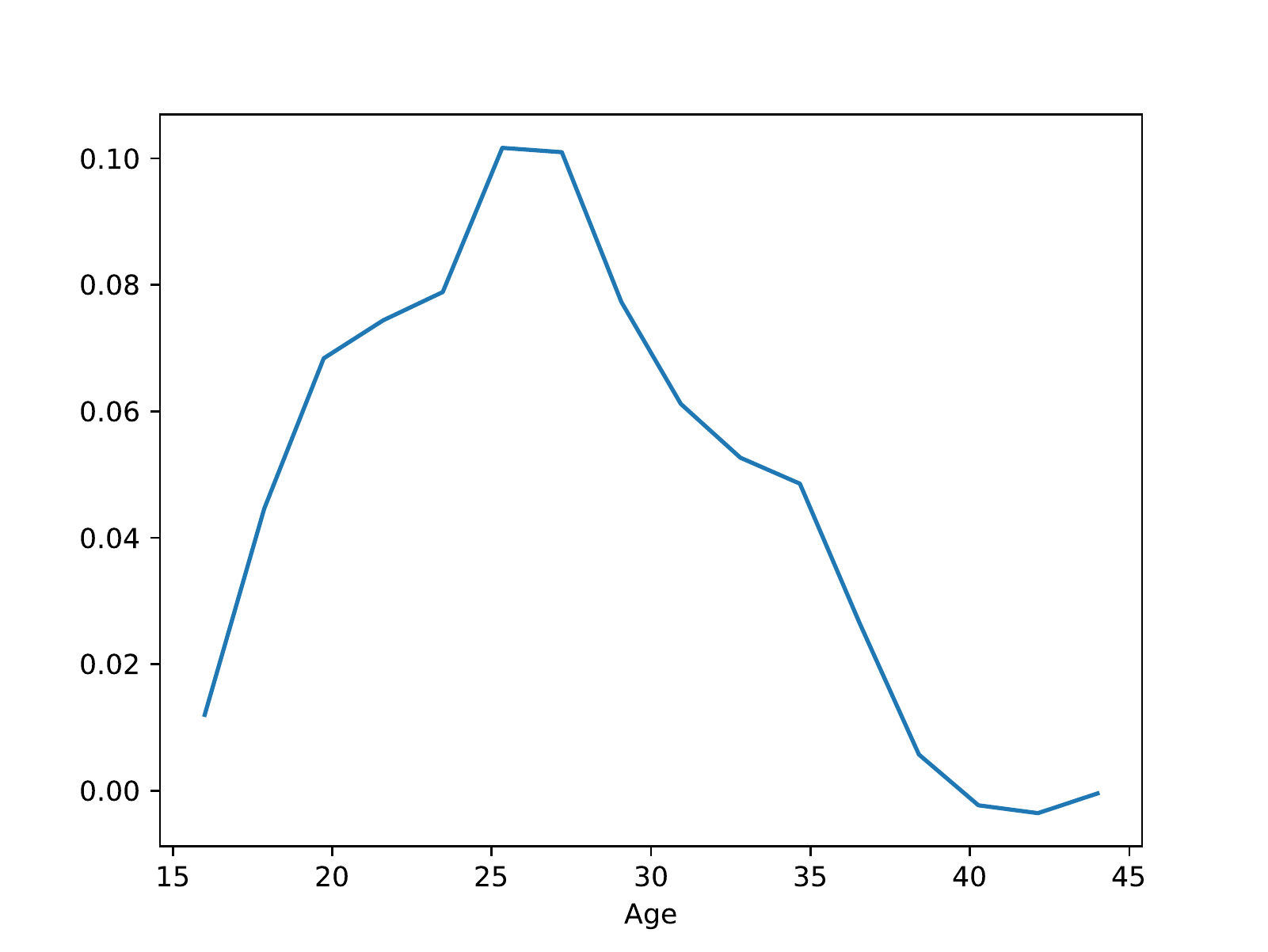}
%\caption{fig2}
\end{minipage}%
}%
\subfigure[$\gamma_j=20$]{
\begin{minipage}[t]{0.3\linewidth}
\centering
\includegraphics[scale=0.15]{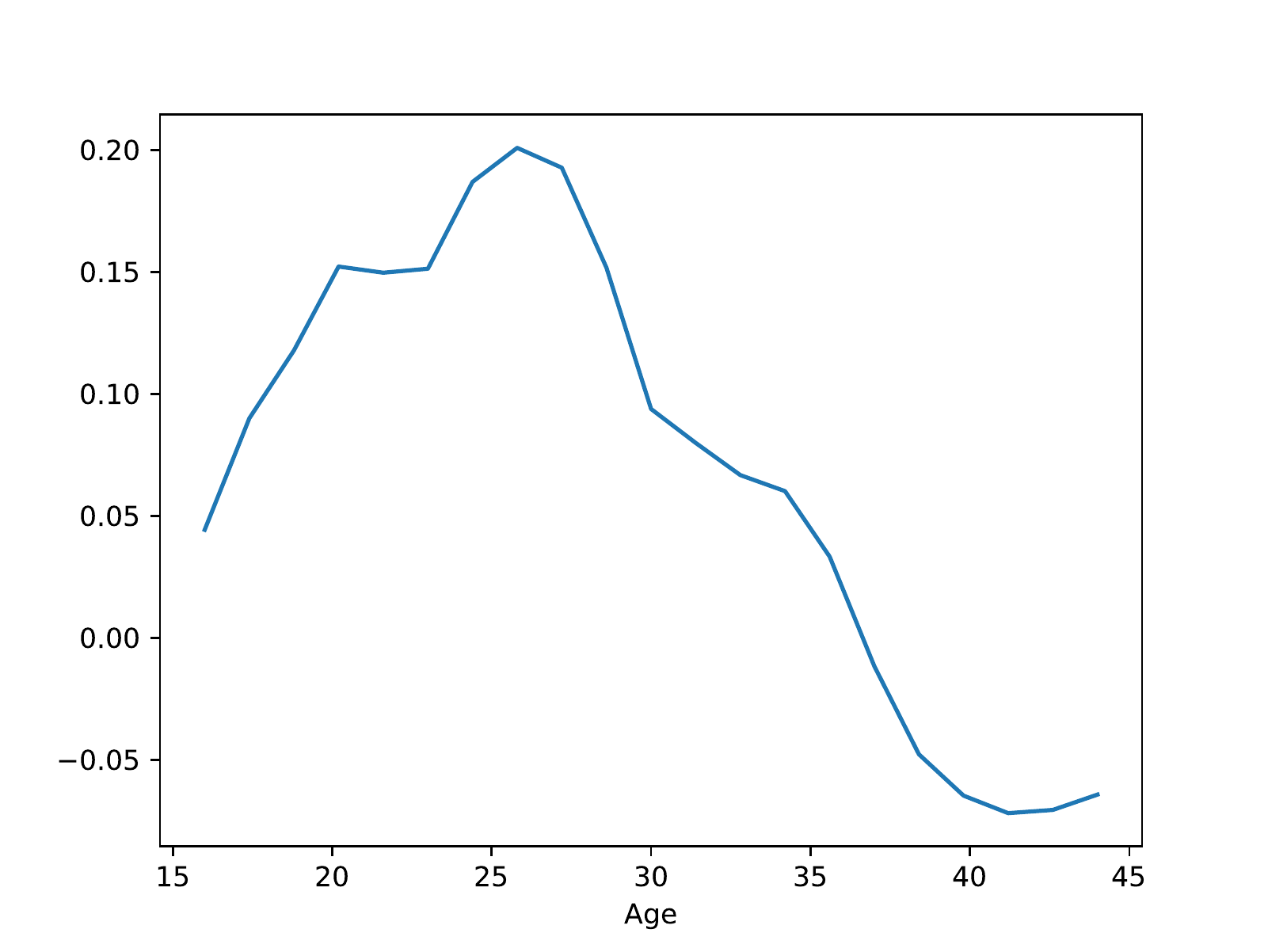}
%\caption{fig2}
\end{minipage}%
}%
\caption{The visualization of learned feature shapes (Age) in dataset Skill by \textit{PiLiD}.}
\label{fig-age}
\end{figure}

To demonstrate the interpretability of \textit{PiLiD} and \textit{PiLiB}, we present feature shapes of \textit{Age} in dataset Skill given different $\gamma_j$. First, from Figure \ref{fig-age}(a) and other plots, we stress the importance of providing feature shapes rather than only feature attributions that uses a single value to describe the feature contribution. All other plots ($\gamma_j\ge 1$) can capture the changing tendency of the relationship between \textit{Age} and players' skill estimation (prediction). Such tendency indicates that, at first, the aging process positively affects players' skill because they can gain experience over time, but it changes when players are older because their response speed is affected. Second, as stated in Schwab and Karlen \shortcite{schwab2019cxplain}, the explanations should be robust. From plots (b) to (e) in Figure \ref{fig-age}, the obtained feature shapes have a similar tendency given predefined number of intervals. That is because the proposed initialization enforces the optimization to start from a more promising region, and thus makes the interpretability more robust and stable. 

\begin{figure}
\centering
\subfigure[Interaction effects between average length of uninterrupted sequences of capital letters and word frequency of \textit{receive}.]{
\begin{minipage}[t]{0.5\linewidth}
\centering
\includegraphics[scale=0.3]{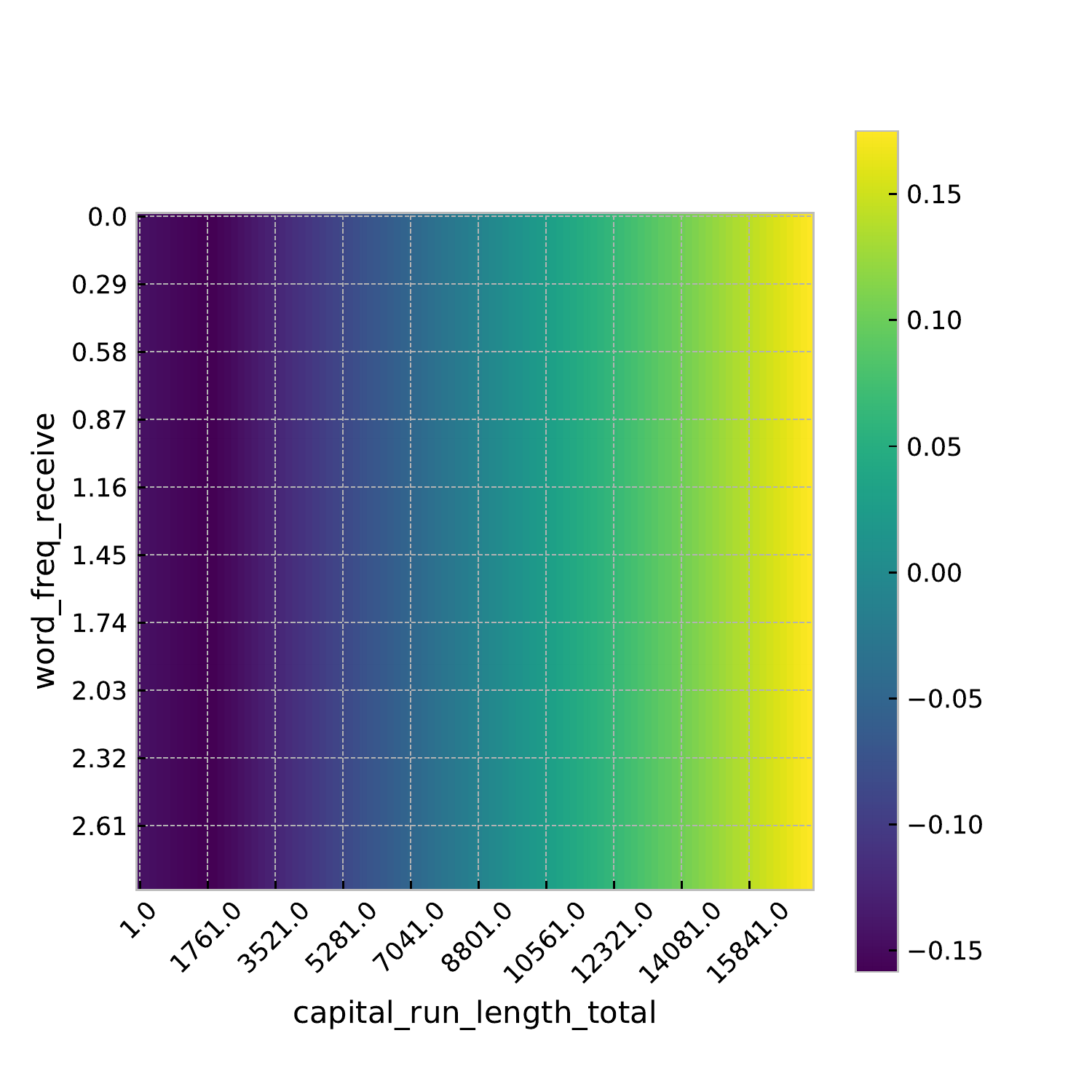}
%\caption{fig1}
\end{minipage}%
}\;%
\subfigure[Interaction effects between word frequency of \textit{pm} and \textit{free}.]{
\begin{minipage}[t]{0.5\linewidth}
\centering
\includegraphics[scale=0.3]{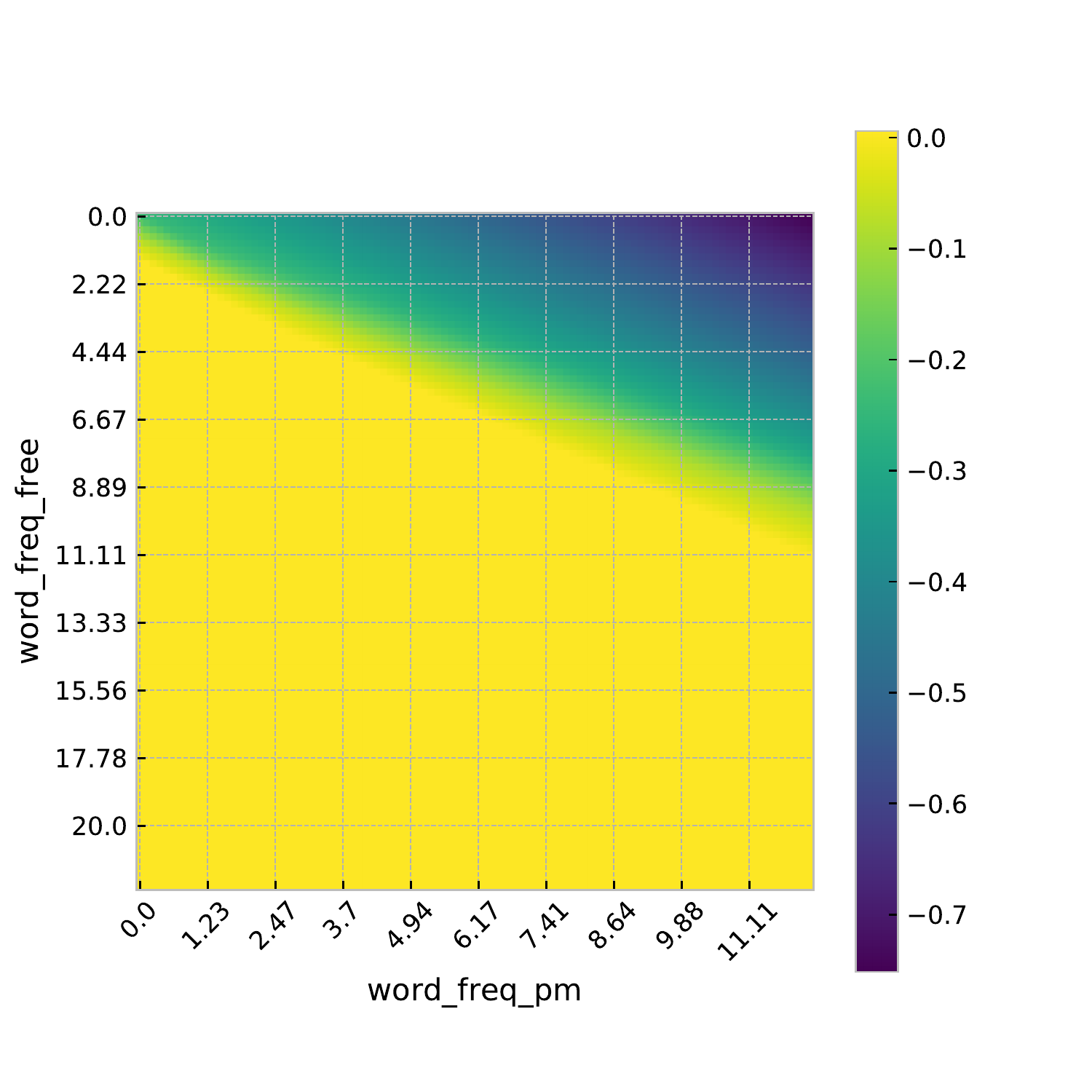}
%\caption{fig2}
\end{minipage}%
}%
\caption{The visualization of interacting features in dataset Spambase by \textit{PiLiB}.}
\label{fig-inter}
\end{figure}

In Figure \ref{fig-inter}, we present an example of the learned interacting effects by \textit{PiLiB}. These relationships help to understand why an email is classified as spam. For example, an email is more likely to be a spam if we observe the overuse of continuous capital letters and less use of the term \textit{receive}. Interpreting and applying these extracted interactions needs further examinations, and will be our future work. 

\section{Conclusion}

We propose a hybrid interpretable model that uses piecewise linear functions to approximate the individual feature contributions. It is flexible to adapt for other model structures. The experiments demonstrate that the model is explicit enough for users to understand and also has state-of-the-art prediction performance. This research shed new light on the joint learning for interpretability and predictability, and the feasibility of using the learned interpretability to enhance the prediction performance.

\section{Supplementary Material}
\subsection{Algorithm for Training \textit{PiLiB}}

Given loss function $\mathcal{R}_{PiLiB}$, the training process of \textit{PiLiB} is in two phases. At the first stage, we jointly train the linear component with parameter regularization $\Omega(\cdot)$ and deep component with only $L_0$ regularization. After the maximum interaction order in the deep component is smaller or equal to the allowable $K$, we go to next phase. In the second phase, the trained `masks' for the first layer are fixed and on this basis, we optimize all parameters in the model with standard regularization. The training process can be summarized by Algorithm \ref{alg-trainprocess2}.

\begin{algorithm*}[t]
\caption{Training process for \textit{PiLiB}.}
\label{alg-trainprocess2}
\begin{algorithmic}[1]
\REQUIRE Training data $\{\textbf{x}_{(i)}, y_i\}^N_{i=1}$, structure of deep component $L$, predefined number of sub-intervals $\gamma_j,j=1,\dots,m$, type of loss function $\mathcal{L}(\cdot)$, regularization term $\lambda$, $\lambda_0$ and regularization type $\Omega(\cdot)$, maximum interaction order $K$, number of blocks $B$, initialization coefficient $\sigma$ for deep component, number of epoch and Batch size.\\
\ENSURE Parameters ${\boldsymbol{\omega} ^*},{\textbf{w}^*},{\textbf{b}^*},{\boldsymbol{\theta} ^*},\textbf{G}^* $. \\
\STATE Steps 1 and 2 in Algorithm 1 in main paper.\\
\WHILE{$\mathop {\max }\limits_{i\in\{1,\ldots,B\}} {\hat{k}_i} \ge K$} 
\STATE Optimize loss function: $\mathcal{R}_1=\mathcal{R}_{PiLiB} + \lambda\Omega\left( \boldsymbol{\omega},{\textbf{w}},{\textbf{b}}\right)$.\\
\STATE ${\boldsymbol{\omega} ^*},{\textbf{w}^*},{\textbf{b}^*},\textbf{G}^*,{\boldsymbol{\theta} ^*} = \mathop {\arg \min }\limits_{\boldsymbol{\omega} ,\textbf{w},\textbf{b}, \textbf{G},\boldsymbol{\theta}} \{ \mathcal{R}_1({\boldsymbol{\omega} ,\textbf{w},\textbf{b}, \textbf{G},\boldsymbol{\theta}})\}$.
\ENDWHILE 
\STATE Fix $\textbf{G}^*$, and initialize the parameters $\{{\boldsymbol{\omega} },{\textbf{w}},{\textbf{b}},{\boldsymbol{\theta} }\}\leftarrow\{{\boldsymbol{\omega} ^*},{\textbf{w}^*},{\textbf{b}^*},{\boldsymbol{\theta} ^*}\}$.\\
\STATE Optimize $\mathcal{R}_2 = \frac{1}{N}\left( {\sum\limits_{i = 1}^N {\mathcal{L}\left( {h\left( {\textbf{x}_{(i)} ;{\boldsymbol{\omega} ,\textbf{w},\textbf{b},{\tilde{\textbf{W}}^1 \odot \mathcal{T}\left( \textbf{G}^*\right) } } } \right), \{\textbf{W}^l\}_{l=2}^{L+1}, \{\textbf{b}^l \}_{l=2}^{L+1},{y_i}} \right)} } \right)+\lambda\Omega\left( \boldsymbol{\omega} ,\textbf{w},\textbf{b}, \boldsymbol\theta \right)$ by standard mini-batch stochastic gradient. \\
\STATE ${\boldsymbol{\omega} ^*},{\textbf{w}^*},{\textbf{b}^*},{\boldsymbol{\theta} ^*} = \mathop {\arg \min }\limits_{\boldsymbol{\omega} ,\textbf{w},\textbf{b}, \boldsymbol{\theta}} \{ \mathcal{R}_2({\boldsymbol{\omega} ,\textbf{w},\textbf{b}, \boldsymbol{\theta}})\}$.
\end{algorithmic}
\end{algorithm*}

\subsection{Extend Simulation Results}

Here we present full simulation results of regression and classification problems. Tables \ref{tab-n5000} to \ref{tab-clas15000} present the simulation results given different numbers of objects and features for two problems. We emphasize three interesting patterns:
\begin{itemize}
\item As stated in main paper, when the dataset is smaller, for example $N=5000$, since the proposed \textit{PiLiD} is under a deep learning scheme, it requires sufficient data to train parameters. Therefore, \textit{PiLiD} and MLP's performance on simulation data are worse than some traditional predictive models. However, when the datasets are larger ($N=10000, 15000, 20000$), the \textit{PiLiD} perform better. 
\item In classification problems, the performance of the proposed \textit{PiLiD} is on a par with or comparable to MLP. The proposed \textit{PiLiD} performs better than MLP given same regression problem settings. That is because the pretriaing process in Algorithm 1 helps \textit{PiLiD} find better solutions. Moreover, the piecewise linear form also increases the expressiveness of the entire model, thereby being more powerful than a pure MLP. 
\item There is a tendency that the predictive performance increases along with the increase of predefined number of intervals. It makes sense because the larger number of intervals, the more complex the linear component. As stated in main paper, an extreme simulation is let $\gamma_j\rightarrow \infty$, then the linear component becomes a GAM, which is more powerful than linear models. However, larger number of intervals will require more computational cost and data samples, as a trade-off, we suggest using $\gamma_j=5, 10, 15$.
\end{itemize}

\begin{table*}
\footnotesize
  \centering
  \caption{Average $MSE$ for regression problems when $N=5000$ (The best results are marked in bold. The number following \textit{PiLiD} is the predefined number of intervals for piecewise linear functions.).}
    \begin{tabular}{lrrrrr}
    \toprule
       \# features   & {10} & {20} & {30} & {40} & {50} \\
          \midrule
    \textit{PiLiD-1} & 0.3002 $\pm$ 0.0258 & 0.2894 $\pm$ 0.0155 & 0.4217 $\pm$ 0.0377 & 0.4264 $\pm$ 0.0230 & 0.4397 $\pm$ 0.0417 \\
    \textit{PiLiD-5} & 0.2976 $\pm$ 0.0162 & 0.2920 $\pm$ 0.0147 & 0.4158 $\pm$ 0.0387 & 0.4252 $\pm$ 0.0260 & 0.4214 $\pm$ 0.0237 \\
    \textit{PiLiD-10} & 0.2993 $\pm$ 0.0256 & 0.2906 $\pm$ 0.0152 & 0.4206 $\pm$ 0.0458 & 0.4238 $\pm$ 0.0328 & 0.4415 $\pm$ 0.0440 \\
    \textit{PiLiD-15} & 0.2955 $\pm$ 0.0341 & 0.2896 $\pm$ 0.0150 &\textbf{ 0.4056 $\pm$ 0.0349} & 0.4196 $\pm$ 0.0246 & 0.4316 $\pm$ 0.0434 \\
    \textit{PiLiD-20} & 0.2954 $\pm$ 0.0262 & 0.2917 $\pm$ 0.0180 & 0.4133 $\pm$ 0.0539 & 0.4163 $\pm$ 0.0250 & 0.4262 $\pm$ 0.0314 \\
    MLP   & 0.3108 $\pm$ 0.0402 & 0.3122 $\pm$ 0.0223 & 0.4623 $\pm$ 0.0418 & 0.4782 $\pm$ 0.0643 & 0.4832 $\pm$ 0.0563 \\
    SVM (rbf kernel)   & 0.6951 $\pm$ 0.0675 & \textbf{0.2803 $\pm$ 0.0139} & 0.4649 $\pm$ 0.0309 & \textbf{0.3759 $\pm$ 0.0168} & \textbf{0.3935 $\pm$ 0.0202} \\
    Random Forest    & \textbf{0.2742 $\pm$ 0.0119} & 0.3330 $\pm$ 0.0152 & 0.4217 $\pm$ 0.0251 & 0.5448 $\pm$ 0.0212 & 0.6183 $\pm$ 0.0264 \\
    Linear Regression    & 0.7195 $\pm$ 0.0545 & 0.4768 $\pm$ 0.0150 & 0.6327 $\pm$ 0.0371 & 0.6167 $\pm$ 0.0392 & 0.6370 $\pm$ 0.0216 \\
    GAM   & 0.3302 $\pm$ 0.0301 & 0.3821 $\pm$ 0.0150 & 0.4767 $\pm$ 0.0309 & 0.4866 $\pm$ 0.0249 & 0.4984 $\pm$ 0.0320 \\
    \bottomrule
    \end{tabular}%
  \label{tab-n5000}%
\end{table*}%

\begin{table*}
\footnotesize
  \centering
  \caption{Average $MSE$ for regression problems when $N=10000$.}
    \begin{tabular}{lrrrrr}
    \toprule
          \# features   & {10} & {20} & {30} & {40} & {50} \\
          \midrule
    \textit{PiLiD-1} & 0.2839 $\pm$ 0.0179 & 0.2940 $\pm$ 0.0142 & 0.3376 $\pm$ 0.0134 & 0.3420 $\pm$ 0.0245 & 0.4561 $\pm$ 0.0344 \\
    \textit{PiLiD-5} & 0.2795 $\pm$ 0.0098 & 0.3035 $\pm$ 0.0267 & 0.3366 $\pm$ 0.0253 & 0.3361 $\pm$ 0.0110 & 0.4478 $\pm$ 0.0434 \\
    \textit{PiLiD-10} & 0.2818 $\pm$ 0.0145 & \textbf{0.2904 $\pm$ 0.0106} & 0.3310 $\pm$ 0.0148 & 0.3393 $\pm$ 0.0197 & \textbf{0.4314 $\pm$ 0.0185} \\
    \textit{PiLiD-15} & 0.2771 $\pm$ 0.0066 & 0.2942 $\pm$ 0.0146 & 0.3386 $\pm$ 0.0275 & \textbf{0.3340 $\pm$ 0.0167} & 0.4357 $\pm$ 0.0321 \\
    \textit{PiLiD-20} & \textbf{0.2770 $\pm$ 0.0087} & 0.2952 $\pm$ 0.0137 & 0.3279 $\pm$ 0.0121 & 0.3384 $\pm$ 0.0219 & 0.4321 $\pm$ 0.0598 \\
    MLP   & 0.2846 $\pm$ 0.0171 & 0.3036 $\pm$ 0.0175 & 0.3419 $\pm$ 0.0213 & 0.3627 $\pm$ 0.0141 & 0.4799 $\pm$ 0.0441 \\
    SVM (rbf kernel)  & 0.2973 $\pm$ 0.0066 & 0.4005 $\pm$ 0.0144 & \textbf{0.3140 $\pm$ 0.0069 }& 0.3815 $\pm$ 0.0111 & 0.4877 $\pm$ 0.0126 \\
    Random Forest    & 0.2710 $\pm$ 0.0060 & 0.3129 $\pm$ 0.0059 & 0.3891 $\pm$ 0.0104 & 0.3923 $\pm$ 0.0092 & 0.5996 $\pm$ 0.0168 \\
    Linear Rregression    & 0.3458 $\pm$ 0.0066 & 0.5705 $\pm$ 0.0201 & 0.3718 $\pm$ 0.0092 & 0.3834 $\pm$ 0.0111 & 0.4894 $\pm$ 0.0121 \\
    GAM   & 0.3240 $\pm$ 0.0043 & 0.4106 $\pm$ 0.0233 & 0.3570 $\pm$ 0.0064 & 0.3679 $\pm$ 0.0129 & 0.4755 $\pm$ 0.0203 \\
    \bottomrule
    \end{tabular}%
  \label{tab-n10000}%
\end{table*}%

\begin{table*}
\footnotesize
  \centering
  \caption{Average $MSE$ for regression problems when $N=15000$.}
    \begin{tabular}{lrrrrr}
          \toprule
          \# features   & {10} & {20} & {30} & {40} & {50} \\
          \midrule
    \textit{PiLiD-1} & 0.2593 $\pm$ 0.0115 & \textbf{0.2972 $\pm$ 0.0087} & 0.3496 $\pm$ 0.0539 & 0.3380 $\pm$ 0.0170 & 0.3470 $\pm$ 0.0295 \\
    \textit{PiLiD-5} & 0.2570 $\pm$ 0.0069 & 0.2999 $\pm$ 0.0115 & 0.3312 $\pm$ 0.0141 & 0.3412 $\pm$ 0.0274 & 0.3439 $\pm$ 0.0203 \\
    \textit{PiLiD-10} & \textbf{0.2567 $\pm$ 0.0071} & 0.3017 $\pm$ 0.0108 & \textbf{0.3214 $\pm$ 0.0111} & 0.3309 $\pm$ 0.0127 & 0.3398 $\pm$ 0.0122 \\
    \textit{PiLiD-15} & 0.2569 $\pm$ 0.0076 & 0.2974 $\pm$ 0.0100 & 0.3226 $\pm$ 0.0124 & 0.3386 $\pm$ 0.0270 & 0.3370 $\pm$ 0.0259 \\
    \textit{PiLiD-20} & 0.2592 $\pm$ 0.0084 & 0.2963 $\pm$ 0.0116 & 0.3215 $\pm$ 0.0116 & \textbf{0.3296 $\pm$ 0.0199} & \textbf{0.3367 $\pm$ 0.0156} \\
    MLP   & 0.2592 $\pm$ 0.0064 & 0.3065 $\pm$ 0.0092 & 0.3463 $\pm$ 0.0172 & 0.3411 $\pm$ 0.0164 & 0.3500 $\pm$ 0.0194 \\
    SVM (rbf kernel)  & 0.2598 $\pm$ 0.0071 & 0.3034 $\pm$ 0.0065 & 0.3325 $\pm$ 0.0076 & 0.3426 $\pm$ 0.0077 & 0.3520 $\pm$ 0.0077 \\
    Random Forest    & 0.2619 $\pm$ 0.0096 & 0.3079 $\pm$ 0.0068 & 0.3357 $\pm$ 0.0072 & 0.4424 $\pm$ 0.0116 & 0.4146 $\pm$ 0.0114 \\
    Linear Regression    & 0.3006 $\pm$ 0.0080 & 0.3605 $\pm$ 0.0077 & 0.4165 $\pm$ 0.0097 & 0.4664 $\pm$ 0.0095 & 0.4574 $\pm$ 0.0088 \\
    GAM   & 0.2790 $\pm$ 0.0072 & 0.3301 $\pm$ 0.0070 & 0.3715 $\pm$ 0.0088 & 0.4069 $\pm$ 0.0096 & 0.3984 $\pm$ 0.0088 \\
    \bottomrule
    \end{tabular}%
  \label{tab-15000}%
\end{table*}%

\begin{table*}
  \footnotesize
  \centering
  \caption{Average $AUC$ for classification problems when $N=5000$.}
    \begin{tabular}{lrrrrr}
          \toprule
          \# features   & {10} & {20} & {30} & {40} & {50} \\
          \midrule
    \textit{PiWiLiD-1} & 0.8405 $\pm$ 0.0018 & 0.9079 $\pm$ 0.0041 & 0.9241 $\pm$ 0.0020 & 0.8870 $\pm$ 0.0018 & 0.9137 $\pm$ 0.0068 \\
    \textit{PiWiLiD-5} & 0.8439 $\pm$ 0.0019 & \textbf{0.9093 $\pm$ 0.0039} & 0.9241 $\pm$ 0.0022 & 0.8854 $\pm$ 0.0017 & 0.9128 $\pm$ 0.0073 \\
    \textit{PiWiLiD-10} & 0.8441 $\pm$ 0.0018 & 0.9076 $\pm$ 0.0040 & 0.9226 $\pm$ 0.0024 & 0.8870 $\pm$ 0.0023 & 0.9132 $\pm$ 0.0070 \\
    \textit{PiWiLiD-15} & 0.8432 $\pm$ 0.0017 & 0.9075 $\pm$ 0.0043 & 0.9244 $\pm$ 0.0025 & 0.8857 $\pm$ 0.0019 & 0.9136 $\pm$ 0.0077 \\
    \textit{PiWiLiD-20} & 0.8455 $\pm$ 0.0019 & 0.9085 $\pm$ 0.0040 & 0.9221 $\pm$ 0.0024 & 0.8872 $\pm$ 0.0018 & 0.9142 $\pm$ 0.0071 \\
    MLP   & 0.8428 $\pm$ 0.0043 & 0.9071 $\pm$ 0.0036 & 0.9225 $\pm$ 0.0056 & 0.8867 $\pm$ 0.0029 & \textbf{0.9146 $\pm$ 0.0065 }\\
    SVM (rbf kernel)  & 0.8029 $\pm$ 0.0223 & 0.8357 $\pm$ 0.0078 & 0.8943 $\pm$ 0.0014 & 0.8468 $\pm$ 0.0016 & 0.8637 $\pm$ 0.0013 \\
    Random Forest    & \textbf{0.8503 $\pm$ 0.0036} & 0.8997 $\pm$ 0.0094 & \textbf{0.9264 $\pm$ 0.0064} & \textbf{0.8929 $\pm$ 0.0056} & 0.9088 $\pm$ 0.0054 \\
    Logistic Regression    & 0.7921 $\pm$ 0.0013 & 0.8074 $\pm$ 0.0055 & 0.8633 $\pm$ 0.0021 & 0.7989 $\pm$ 0.0019 & 0.8215 $\pm$ 0.0015 \\
    GAM   & 0.8398 $\pm$ 0.0014 & 0.8867 $\pm$ 0.0028 & 0.9119 $\pm$ 0.0036 & 0.8386 $\pm$ 0.0025 & 0.8953 $\pm$ 0.0034 \\
    \bottomrule
    \end{tabular}%
  \label{tab-cla5000}%
\end{table*}%

\begin{table*}
  \footnotesize
  \centering
  \caption{Average $AUC$ for classification problems when $N=10000$.}
    \begin{tabular}{lrrrrr}
          \toprule
          \# features   & {10} & {20} & {30} & {40} & {50} \\
          \midrule
    \textit{PiWiLiD-1} & 0.8401 $\pm$ 0.0077 & 0.9458 $\pm$ 0.0044 & 0.9271 $\pm$ 0.0056 & 0.9116 $\pm$ 0.0713 & 0.9404 $\pm$ 0.0073 \\
    \textit{PiWiLiD-5} & 0.8402 $\pm$ 0.0079 & 0.9454 $\pm$ 0.0043 & 0.9272 $\pm$ 0.0052 & 0.9085 $\pm$ 0.0649 & 0.9403 $\pm$ 0.0073 \\
    \textit{PiWiLiD-10} & 0.8401 $\pm$ 0.0079 & 0.9456 $\pm$ 0.0042 & 0.9271 $\pm$ 0.0054 & \textbf{0.9136 $\pm$ 0.0473 }& 0.9408 $\pm$ 0.0072 \\
    \textit{PiWiLiD-15} & 0.8399 $\pm$ 0.0078 & 0.9479 $\pm$ 0.0040 & 0.9275 $\pm$ 0.0055 & 0.9071 $\pm$ 0.0594 & 0.9409 $\pm$ 0.0074 \\
    \textit{PiWiLiD-20} & 0.8412 $\pm$ 0.0082 & \textbf{0.9481 $\pm$ 0.0040} & \textbf{0.9276 $\pm$ 0.0054 }& 0.9128 $\pm$ 0.0458 & 0.9410 $\pm$ 0.0071 \\
    MLP   & \textbf{0.8428 $\pm$ 0.0080} & 0.9463 $\pm$ 0.0041 & 0.9265 $\pm$ 0.0055 & 0.9107 $\pm$ 0.0679 & \textbf{0.9418 $\pm$ 0.0079} \\
    SVM (rbf kernel)  & 0.7229 $\pm$ 0.0112 & 0.8526 $\pm$ 0.0078 & 0.8094 $\pm$ 0.0101 & 0.7940 $\pm$ 0.0646 & 0.8874 $\pm$ 0.0023 \\
    Random Forest    & 0.7403 $\pm$ 0.0099 & 0.8297 $\pm$ 0.0067 & 0.8738 $\pm$ 0.0071 & 0.8653 $\pm$ 0.0462 & 0.8588 $\pm$ 0.0094 \\
    Logistic Regression    & 0.7866 $\pm$ 0.0113 & 0.8044 $\pm$ 0.0065 & 0.6263 $\pm$ 0.0162 & 0.6953 $\pm$ 0.0570 & 0.8195 $\pm$ 0.0035 \\
    GAM   & 0.7689 $\pm$ 0.0074 & 0.8750 $\pm$ 0.0058 & 0.9175 $\pm$ 0.0046 & 0.8756 $\pm$ 0.0652 & 0.9212 $\pm$ 0.0074 \\
    \bottomrule
    \end{tabular}%
  \label{tab-cla10000}%
\end{table*}%

\begin{table*}
  \footnotesize
  \centering
  \caption{Average $AUC$ for classification problems when $N=15000$.}
    \begin{tabular}{lrrrrr}
          \toprule
          \# features   & {10} & {20} & {30} & {40} & {50} \\
          \midrule
    \textit{PiWiLiD-1} & 0.7281 $\pm$ 0.0061 & 0.9064 $\pm$ 0.0027 & 0.9057 $\pm$ 0.0042 & 0.9406 $\pm$ 0.0028 & 0.9206 $\pm$ 0.0047 \\
    \textit{PiWiLiD-5} & 0.7307 $\pm$ 0.0061 & 0.9065 $\pm$ 0.0027 & 0.9059 $\pm$ 0.0042 & 0.9411 $\pm$ 0.0031 & 0.9205 $\pm$ 0.0056 \\
    \textit{PiWiLiD-10} & 0.7320 $\pm$ 0.0071 & 0.9066 $\pm$ 0.0028 & 0.9064 $\pm$ 0.0039 & 0.9406 $\pm$ 0.0030 & 0.9209 $\pm$ 0.0054 \\
    \textit{PiWiLiD-15} & 0.7329 $\pm$ 0.0070 & 0.9066 $\pm$ 0.0029 & 0.9067 $\pm$ 0.0042 & 0.9411 $\pm$ 0.0031 & 0.9212 $\pm$ 0.0074 \\
    \textit{PiWiLiD-20} & \textbf{0.7337 $\pm$ 0.0067} & 0.9068 $\pm$ 0.0029 & \textbf{0.9081 $\pm$ 0.0040} & \textbf{0.9420 $\pm$ 0.0030} & \textbf{0.9215 $\pm$ 0.0071} \\
    MLP   & 0.7297 $\pm$ 0.0057 & \textbf{0.9086 $\pm$ 0.0032} & 0.9078 $\pm$ 0.0038 & 0.9417 $\pm$ 0.0032 & 0.9211 $\pm$ 0.0048 \\
    SVM (rbf kernel)  & 0.7077 $\pm$ 0.0086 & 0.7782 $\pm$ 0.0042 & 0.8094 $\pm$ 0.0101 & 0.8007 $\pm$ 0.0066 & 0.8874 $\pm$ 0.0023 \\
    Random Forest    & 0.7103 $\pm$ 0.0086 & 0.7790 $\pm$ 0.0080 & 0.7484 $\pm$ 0.0074 & 0.8043 $\pm$ 0.0102 & 0.8588 $\pm$ 0.0094 \\
    Logistic Regression    & 0.6082 $\pm$ 0.0049 & 0.7579 $\pm$ 0.0030 & 0.7138 $\pm$ 0.0043 & 0.7981 $\pm$ 0.0049 & 0.8096 $\pm$ 0.0086 \\
    GAM   & 0.6808 $\pm$ 0.0072 & 0.8349 $\pm$ 0.0041 & 0.8520 $\pm$ 0.0043 & 0.8854 $\pm$ 0.0047 & 0.8644 $\pm$ 0.0084 \\
    \bottomrule
    \end{tabular}%
  \label{tab-clas15000}%
\end{table*}%

\subsection{Real-world Datasets}

The datasets used in main paper can be downloaded from the following websites:
\begin{itemize}
\item Bike sharing: \url{https://archive.ics.uci.edu/ml/datasets/Bike+Sharing+Dataset}
\item Bank marketing: \url{https://archive.ics.uci.edu/ml/datasets/Bank+Marketing}
\item Spambase: \url{https://archive.ics.uci.edu/ml/datasets/Spambase}
\item Skill: \url{http://archive.ics.uci.edu/ml/datasets/SkillCraft1+Master+Table+Dataset}
\end{itemize}

\bibliographystyle{named}
\bibliography{ijcai20}

\end{document}